\documentclass{article}

\PassOptionsToPackage{numbers, compress}{natbib}

\usepackage[preprint]{neurips_2026}

\usepackage[utf8]{inputenc}
\usepackage[T1]{fontenc}
\usepackage{hyperref}
\usepackage{url}
\usepackage{booktabs}
\usepackage{amsfonts}
\usepackage{amsmath}
\usepackage{amssymb}
\usepackage{nicefrac}
\usepackage{microtype}
\usepackage[table]{xcolor}
\usepackage{graphicx}

\title{Colinearity Decay: Training Quantization-Friendly\\  ViTs with Outlier Decay}

\author{
  Jin Tong$^{1,2}$ \qquad
  Guang Liang$^{1,2,3}$ \qquad
  Peilin Sun$^{1,2}$ \qquad
  Jianxin Wu$^{1,2*}$ \\
  $^{1}$State Key Laboratory of Novel Software Technology, Nanjing University, Nanjing 210023, China \\
  $^{2}$School of Artificial Intelligence, Nanjing University, Nanjing 210023, China \\
  $^{3}$Zhongguancun Academy, Beijing 100094, China \\
  \texttt{\{tongj, liangg, sunpl\}@lamda.nju.edu.cn, wujx2001@nju.edu.cn}
}

\begin{document}

\maketitle

\begin{abstract}
Low-bit quantization is a practical route for efficiently deploying vision Transformers, yet activation outliers complicate fully quantized deployment. Existing methods either handle quantization post-training or suppress large activations during training; however, aggressively restricting outliers in vision models can lead to a poorer trade-off between full-precision and quantized accuracy. We argue that rather than simply suppressing outliers, the training objective should control the structural amplification that makes them harmful. To this end, we introduce Colinearity-Decay (CD), a structural regularizer for ordered matrix pairs within Transformer blocks. CD penalizes detrimental cross-matrix alignment and mitigates extreme activations without altering the architecture or task loss. Applied as a decoupled update, CD is non-invasive and introduces minimal training overhead. Across ImageNet-1K pre-training, COCO detection, and downstream fine-tuning, CD consistently boosts quantized accuracy across multiple pipelines while preserving, or even improving, full-precision performance. Ultimately, our results demonstrate that structural regularization effectively prepares vision Transformers for low-bit deployment with zero inference-time overhead.
\end{abstract}

\section{Introduction}

Vision Transformers~\citep{vaswani2017transformer,dosovitskiy2020vit,liu2021swin,touvron2021deit} have become standard backbones for image classification, detection, and transfer learning. Their deployment cost, however, remains a practical obstacle when memory bandwidth, latency, or energy is limited. Low-bit quantization is a natural way to reduce this cost, but fully quantizing Transformer is still difficult. The main difficulty often comes from activations: a small number of channels or tokens can take much larger values than the rest, forcing coarse quantization scales and leaving the majority of activations represented with low effective precision.

Existing quantization methods for vision Transformers address this problem mostly at deployment time, through calibration~\citep{yuan2022ptq4vit,lin2021fqvit}, scale reparameterization~\citep{li2023repq,yang2024dopq}, reconstruction~\citep{zhong2025erq,wu2025aphq,wu2025fimaq}, or QAT methods that learn quantization parameters and stabilize low-bit training dynamics~\citep{esser2019lsq,li2022qvit,liu2023ofq,huang2023quantizationvariation,liang2025gplq}. These methods are effective, but mostly applied \emph{after the full-precision model has already been trained}. The model itself may remain poorly conditioned for low-bit quantization.

Another line of work reduces activation outliers \emph{during training}. Architectural or optimization changes can mitigate attention sinks and other sources of extreme activations~\citep{kaul2024softmax1,sun2024massive,agarwal2025attnbias,he2024opb,park2025osp}, and direct activation regularization penalizes large intermediate values~\citep{nrusimha2024actreg,liang2025tweo}. These approaches address outliers through different routes, but they lead to the same \emph{question for quantization-friendly training}: shall we minimize outliers as aggressively as possible, or \emph{shall we only control the part of the outlier structure that harms low-bit deployment}?

In this paper we argue that the second target is more appropriate for vision models. Direct activation-suppression methods such as TWEO~\citep{liang2025tweo} are closely tied to FP8 training: reducing maximum activation values (e.g., to 30) will surely stabilize low-precision optimization on the FP8 hardware. But, these methods will potentially harm the accuracy of the trained FP models, and as a consequence harm the quantized models. 

\begin{figure}[t]
	\centering
	\includegraphics[width=\linewidth]{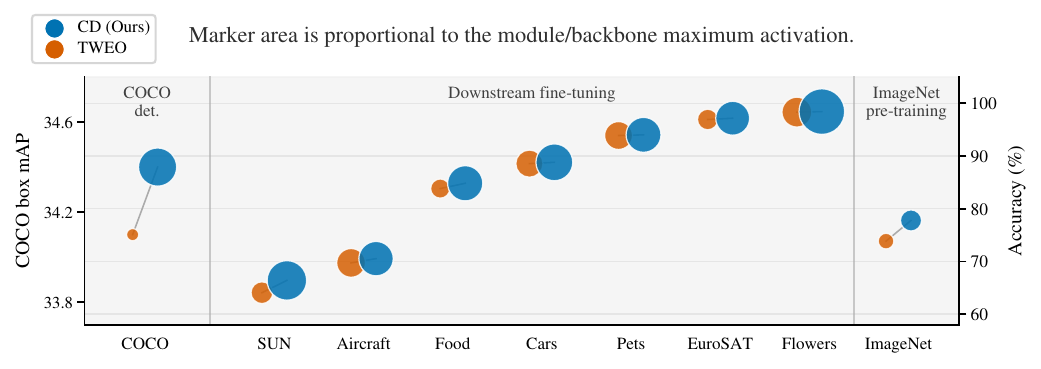}
	\caption{Comparison between CD and TWEO under RepQ-ViT~\citep{li2023repq} W4A4 quantization. CD keeps larger maximum activations yet gives higher quantized accuracy, especially in object detection.}
	\label{fig:intro_motivation}
\end{figure}

Figure~\ref{fig:intro_motivation} illustrates and empirically supports our argument by comparing TWEO with Colinearity-Decay (CD), the method we will propose in this paper. TWEO~\citep{liang2025tweo} strongly suppresses maximum activations (outliers), and \emph{CD often has larger maximum activations}---the blue circles have larger areas than their corresponding orange colored counterparts. But, \emph{CD at the same time achieves higher quantization accuracy}. This does not mean that outliers are harmless. Rather, maximum activation behaves as a diagnostic quantity, instead of a self-sufficient training objective. 

That is, our key argument is that \emph{a quantization-friendly training method should reduce harmful extreme activation values to a reasonable level, but not down to a level that is too small to potentially harm the representation learned by the full-precision model}. 

\begin{figure}[t]
\centering
\includegraphics[width=\linewidth]{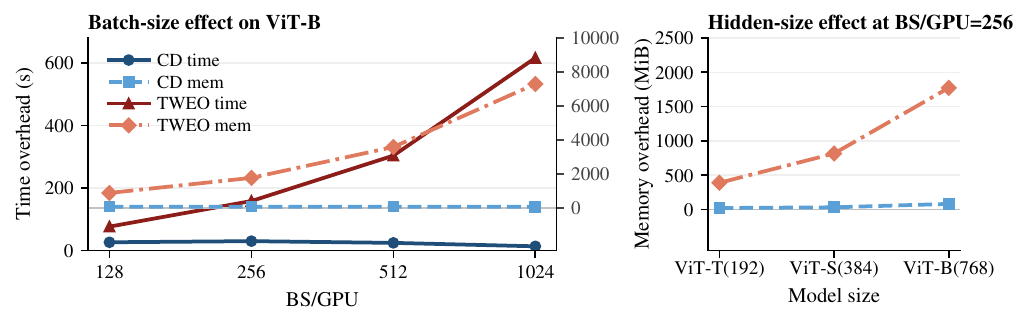}
\caption{Training overhead relative to the baseline run on ImageNet-1K pre-training. Left: time and memory overhead versus per-GPU batch size on ViT-B. Right: memory overhead versus hidden size at BS/GPU$=256$. Time is measured in seconds. Memory is the peak \texttt{max\_allocated} value in MiB.}
\label{fig:exp_efficiency}
\end{figure}

Figure~\ref{fig:exp_efficiency} further reveals an essential practical advantage of our CD method: because CD is applied as a decoupled matrix update (i.e., decay on parameters) rather than as an activation-regularization loss, it adds significantly smaller training overhead than TWEO.

CD is a decoupled regularizer for ordered matrix pairs in Transformer blocks~\citep{vaswani2017transformer}. The method is motivated by a structural source of activation amplification: for a pair of adjacent or functionally adjacent matrices $(W_1, W_2)$, large outputs can arise when rows of $W_2$ align with dominant left singular directions of $W_1$. Standard weight decay regularizes each matrix independently and does not directly penalize this cross-matrix alignment. CD instead penalizes the pairwise energy $\|W_2 W_1\|_F^2$, thus reducing large activation values (i.e., outliers).

This design also leads to a simple implementation. Instead of adding an auxiliary loss to the training graph, CD applies an update in the spirit of decoupled weight decay~\citep{loshchilov2017adamw} to the downstream matrix in each selected pair after the task gradient has been computed. The pairing rule covers both composable pairs, such as normalization scales followed by projections, and functional pairs in attention and FFN sub-layers. As a result, CD keeps the architecture, task loss, and inference \emph{intact}, while adding only a small amount of training-time matrix computation. This non-invasive form also makes CD complementary to architectural outlier-mitigation methods.

We evaluate CD on ImageNet-1K~\citep{deng2009imagenet} pre-training, COCO detection~\citep{lin2014coco} with joint detector training, and downstream classification fine-tuning on nine datasets. Across these settings, CD preserves or slightly improves full-precision accuracy and improves W4A4 quantized accuracy under several quantization pipelines, including percentile PTQ~\citep{banner2019percentile}, RepQ-ViT~\citep{li2023repq}, APHQ~\citep{wu2025aphq}, and GPLQ~\citep{liang2025gplq}. The gains are consistent across ViT and Swin backbones.

Our contributions are as follows:
\begin{itemize}
    \item We empirically find that keeping outliers in a suitable range benefits full-precision vision Transformers and especially their quantized counterparts: huge outliers is harmful, but very small activation values are potentially harmful, too.
    \item We identify excessive cross-matrix alignment as a structural source of activation outliers and introduce Colinearity-Decay, a non-invasive matrix-pair regularizer with negligible overhead. CD is orthogonal to existing quantization and outlier-mitigation methods.
    \item We validate CD across pre-training, downstream fine-tuning, and COCO detection. CD improves quantized accuracy under multiple W4A4 pipelines while preserving full-precision performance. In particular, CD achieves clear gains in the less-studied detection task.
\end{itemize}

\section{Related Work}

\textbf{Quantization Methods for Vision Transformers.} Transformer quantization has been widely studied as a practical route to low-bit deployment, but activation quantization remains difficult because Transformer activations often exhibit strong channel variation and large outliers. For vision Transformers, early PTQ methods~\citep{yuan2022ptq4vit,lin2021fqvit} improve calibration and quantizer design for attention and MLP activations. Later scale-reparameterization or outlier-aware quantization methods~\citep{li2023repq,yang2024dopq} address this difficulty by shifting part of the burden from activations to weights, while reconstruction or error-reduction objectives~\citep{zhong2025erq,wu2025aphq,wu2025fimaq} better match full-precision and quantized models. GPLQ~\citep{liang2025gplq} makes a similar observation from the QAT side, focusing its lightweight adaptation on the harder activation quantization problem. These works show that activation outliers are a central obstacle in Transformer quantization. CD is complementary to this line of work: it trains or fine-tunes the full-precision model to reduce structural sources of large activations, making the resulting model easier to quantize under multiple pipelines.

\textbf{Training-Time Outlier Mitigation.} Another line of work tries to mitigate outliers during training rather than only adapting the quantizer post-training. Some methods~\citep{kaul2024softmax1,sun2024massive,agarwal2025attnbias} connect outliers to attention sinks and introduce architectural changes such as softmax1, kv-bias, or attn-bias to reduce sink behavior or extreme activations; in vision Transformers, register tokens~\citep{darcet2023registers} can also serve as sink tokens. Other work~\citep{qiu2026residualsink,he2024opb,kaul2024softmax1,park2025osp} attributes outliers to residual or normalization structures, block design, or optimization choices, and modifies the architecture or pre-training recipe accordingly. Direct activation-level methods~\citep{nrusimha2024actreg,liang2025tweo} instead penalize or regularize large activations during training, including activation regularization for language-model quantization and TWEO for suppressing extreme Transformer outliers. CD shares the goal of making the trained model more quantization-friendly, but it keeps the architecture and task loss unchanged and applies a decoupled decay to ordered matrix pairs. It therefore targets a structural source of activation outliers while remaining compatible with architectural outlier-mitigation methods.

\textbf{Outliers and Quantization Difficulty.} The relation between outliers and quantization accuracy is more subtle than a single activation-magnitude objective. TWEO~\citep{liang2025tweo} shows that removing extreme activation outliers can enable FP8 training and improve quantization, but our experiments indicate that aggressive activation suppression is not always the best trade-off for quantizing vision models. A related activation-regularization method~\citep{nrusimha2024actreg} notes that directly reducing activation outliers can move quantization difficulty to weights if the regularization is not balanced, while an outlier-suppression study~\citep{wei2022outliersuppression} shows that outliers differ in importance and clipping tolerance. From the optimizer perspective, a related study~\citep{vlassis2025beyondoutliers} finds that common outlier metrics such as max-to-mean ratio and kurtosis do not reliably predict PTQ performance across optimizers. These observations are in line with our view that large activations should be reduced to a suitable range while preserving full-precision accuracy and improving compatibility with downstream quantization methods.

\section{Method}

CD targets a structural source of activation outliers. In a Transformer block~\citep{vaswani2017transformer}, large activation can be amplified not only by an individual matrix, but also by the alignment between two ordered matrices along the data flow. We first formalize this cross-matrix colinearity, then identify the relevant pairs in Transformer blocks, and finally give the decoupled CD update.

\subsection{Colinearity as a Structural Source of Outliers}

Consider an ordered pair of matrices $(W_1, W_2)$, where $W_1 \in \mathbb{R}^{d_m \times d_{\mathrm{in}}}$ and $W_2 \in \mathbb{R}^{d_{\mathrm{out}} \times d_m}$. The notation is local to the pair: $W_1$ is the upstream matrix and $W_2$ is the downstream matrix. 

Let $W_1 = U S V^\top$ be the SVD of $W_1$, where $U = [u_1, \dots, u_r]$ contains the left singular vectors and $S = \mathrm{diag}(s_1, \dots, s_r)$. For the $j$-th row $w_{2,j}^\top$ of $W_2$, the $j$-th coordinate of the composed output is
\begin{equation}
y_j
= w_{2,j}^\top W_1 x
= \sum_{i=1}^{r} s_i \, (w_{2,j}^\top u_i)\, (v_i^\top x)\,.
\label{eq:method_row_decomp}
\end{equation}
This decomposition (from~\citep{liang2025tweo}) isolates the mechanism that CD controls. If a row of $W_2$ aligns with a dominant left singular direction of $W_1$, then the corresponding term is weighted by a large singular value and can amplify selected input components. Such amplification can create large output activation values even when neither matrix is unusual in isolation.

We further \emph{propose a global measure of this pairwise amplification}, which is
\begin{equation}
\|W_2 W_1\|_F^2
= \|W_2 U S\|_F^2\,.
\label{eq:method_pair_energy}
\end{equation}
This quantity is large when rows of $W_2$ place mass on high-singular-value directions of $W_1$. Standard weight decay cannot penalize this cross-matrix alignment.

\subsection{Matrix Pairs in Transformer Blocks}

\begin{figure}[t]
\centering
\includegraphics[width=\linewidth]{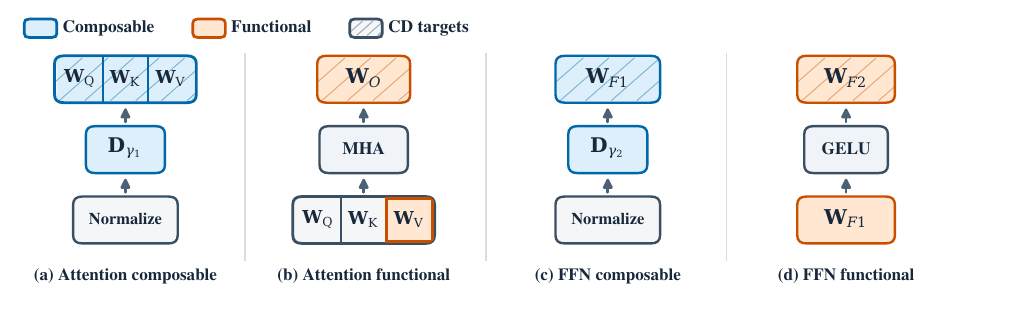}
\caption{Matrix pairs regularized by CD in a pre-norm Transformer block. Blocks with hatching background marks the matrix receiving the decoupled CD update within each pair.}
\label{fig:method_schematic}
\end{figure}

We apply the definition above to the data flow of a standard pre-norm Transformer block~\citep{radford2019gpt2}:
\begin{equation}
X' = X + \mathrm{Attn}(\mathrm{LN}_1(X))\,,
\qquad
Y = X' + \mathrm{FFN}(\mathrm{LN}_2(X'))\,.
\label{eq:method_prenorm_block}
\end{equation}
Figure~\ref{fig:method_schematic} shows the selected pairs. We call an exact local matrix chain a \emph{composable pair}; when two matrices are separated by a nonlinearity or a data-dependent operator but still form a functional route for amplification, we call them a \emph{functional pair}.

\paragraph{Composable pairs.}
The first case comes from normalization scales and the following projections. Omitting the bias term for clarity, LayerNorm or RMSNorm can be written as
\begin{equation}
\mathrm{LN}(X) = \hat{X} D_{\gamma}\,,
\label{eq:method_ln_decomp}
\end{equation}
where $\hat{X}$ is the normalized activation and $D_{\gamma}$ is the diagonal matrix induced by the scale vector $\gamma$. This exposes exact matrix pairs in both attention and FFN branches, shown in Fig.~\ref{fig:method_schematic}(a) and (c):
\begin{equation}
[\,Q\,|\,K\,|\,V\,] = \hat{X}_1 D_{\gamma_1} [W_Q|W_K|W_V]\,,\qquad
H = \hat{X}_2 D_{\gamma_2} W_{F1}\,.
\label{eq:method_composable_examples}
\end{equation}
These pairs can generate outliers by the same mechanism as Eq.~\eqref{eq:method_row_decomp}: a large scale can further amplify the channel associated with a direction, producing a large output along that direction. This view also connects to reparameterization methods in PTQ such as RepQ-ViT~\citep{li2023repq}, which rescale adjacent activation-weight channels to shift quantization difficulty from per-tensor activations to per-channel weights. CD instead reduces this amplification during training.

\paragraph{Functional pairs.}
The second case is not always an exact matrix product, but the relevant matrices still form a route for outlier amplification. When the attention functional pair in Fig.~\ref{fig:method_schematic}(b) simply follows the single-head self-attention. The attention branch can be written as
\begin{equation}
A = \mathrm{softmax}\!\left(\frac{QK^\top}{\sqrt{d_h}}\right)\,,\qquad
\mathrm{Attn}(X) = A V W_O = A X W_V W_O\,.
\label{eq:method_single_head_vo}
\end{equation}
Once $A$ is fixed, $W_V$ and $W_O$ are adjacent linear transforms. Multi-head attention does not merge them into one dense product, but $W_O$ still consumes the directions produced by the value branches. We therefore treat $(W_V, W_O)$ as a functional pair.

The same idea applies to the feed-forward network, giving the FFN functional pair in Fig.~\ref{fig:method_schematic}(d):
\begin{equation}
\mathrm{FFN}(x) = W_{F2}\, \phi(W_{F1} x)\,,
\label{eq:method_ffn_exact}
\end{equation}
where $\phi$ denotes the activation function. Ignoring the nonlinearity gives the diagnostic surrogate
\begin{equation}
\widehat{\mathrm{FFN}}(x) = W_{F2} W_{F1} x\,.
\label{eq:method_ffn_surrogate}
\end{equation}
Note that the colinearity proposed by TWEO~\cite{liang2025tweo} is the one shown in Fig.~\ref{fig:method_schematic}(d). 

Appendix~\ref{app:ffn-outlier-colinearity} shows that some high-energy $W_{F1}$ singular directions align strongly with selected $W_{F2}$ rows, matching Eq.~\eqref{eq:method_row_decomp}. Directly zeroing out the most aligned directions (without retraining) sharply reduces the maximum activation, keeps FP accuracy close to the baseline, and also improves quantized accuracy. This suggests reducing colinearity can be effective in attenuating outliers and improving quantization. 

While TWEO~\cite{liang2025tweo} only considers the colinearity in Fig.~\ref{fig:method_schematic}(d), we \emph{propose new types of colinearity} in Fig.~\ref{fig:method_schematic}(a) to (c), and \emph{propose a unified colinearity decay (CD) method  to handle all of them}.

\subsection{Colinearity-Decay}

For each selected pair, CD penalizes the pairwise energy
\begin{equation}
l_{\mathrm{cd}}(W_1, W_2) = \|W_2 W_1\|_F^2\,.
\label{eq:method_cd_objective}
\end{equation}
The matrix gradients are
\begin{equation}
\frac{\partial l_{\mathrm{cd}}}{\partial W_1}
= 2 W_2^\top W_2 W_1\,,
\qquad
\frac{\partial l_{\mathrm{cd}}}{\partial W_2}
= 2 W_2 W_1 W_1^\top \,.
\label{eq:method_cd_grads}
\end{equation}
Using $W_1=USV^\top$, the row-wise gradient with respect to the downstream matrix is
\begin{equation}
\frac{\partial l_{\mathrm{cd}}}{\partial w_{2,j}^\top}
= 2 w_{2,j}^\top U S^2 U^\top.
\label{eq:method_cd_row_grad}
\end{equation}
Thus, CD shrinks downstream rows along high-singular-value directions of the upstream matrix, which are the directions most responsible for the amplification in Eq.~\eqref{eq:method_row_decomp}. Since $US^2U^\top$ is positive semi-definite, the row-wise CD direction forms a non-obtuse angle with the weight-decay direction, indicating that CD and weight decay are cooperative. See Appendix~\ref{app:cd-details} for these derivations.

In practice, we apply CD only to the second matrix $W_2$ in each local pair. This avoids repeatedly decaying matrices that appear in two local roles: for example, $W_{F1}$ is downstream in $(D_{\gamma_2}, W_{F1})$ and upstream in $(W_{F1}, W_{F2})$. Table~\ref{tab:exp_ablation_pairs} further shows that decaying both matrices in each functional pair gives worse quantized accuracy than decaying only the downstream matrix.

The update is implemented in a decoupled form, similar in spirit to AdamW-style~\citep{loshchilov2017adamw} weight decay. After the task gradient has been computed and before \texttt{optimizer.step()}, we apply
\begin{equation}
W_2 \leftarrow
W_2 - \eta \lambda_{\mathrm{cd}}
W_2 W_1 W_1^\top\,,
\label{eq:method_cd_update}
\end{equation}
where $\eta$ is the learning rate and $\lambda_{\mathrm{cd}}$ is the CD coefficient. The global factor $2$ in Eq.~\eqref{eq:method_cd_grads} is absorbed into $\lambda_{\mathrm{cd}}$. This design keeps the architecture, task loss, and inference graph unchanged.

We apply CD to all Transformer blocks. Composable pairs include the normalization scale with $[W_Q|W_K|W_V]$, and $W_{F1}$. Functional pairs include $W_V \rightarrow W_O$ and $W_{F1} \rightarrow W_{F2}$. In Swin~\citep{liu2021swin}, we include the normalization-scale/reduction pair in downsampling as composable. Gated MLPs~\citep{shazeer2020glu} are not included, although the same pairwise definition can be extended to them. In implementation, we use a normalized CD objective and reduce the weight-decay coefficient under a fixed regularization budget. These details are given in Appendix~\ref{app:cd-details}; they keep the baseline and CD models at roughly comparable per-matrix Frobenius norms, which makes the comparison better controlled.

\section{Experiments}

We evaluate CD on three settings: ImageNet-1K~\citep{deng2009imagenet} pre-training, COCO~\citep{lin2014coco} detection with joint detector training, and downstream image classification fine-tuning. In this part, we mainly want to answer two questions:
\begin{itemize}
	\item Can CD improve low-bit deployment across diverse quantization pipelines while preserving or improving full-precision accuracy? 
	\item Does regularizing the structural alignment that gives rise to activation outliers provide a better overall trade-off than direct activation regularization (as in TWEO~\citep{liang2025tweo}), in terms of training overhead, FP accuracy, and quantized accuracy?
\end{itemize} 

\subsection{Experimental Setup}

\begin{table}[t]
\caption{ImageNet-1K pre-training. TWEO~\cite{liang2025tweo} is reproduced under the same training framework with aligned base hyperparameters and author-provided method-specific hyperparameters. The baseline is a regularly trained model, i.e., without our colinearity decay.}
\label{tab:exp_pretrain_main}
\centering
\footnotesize
\setlength{\tabcolsep}{4pt}
\begin{tabular}{clcccc}
\toprule
Model & Method & FP Top-1 & Max. Act. (mod./blk.) & RepQ-ViT & APHQ \\
\midrule
\smash{\raisebox{-1.0\normalbaselineskip}{ViT-B}} & Baseline & 81.444 & \phantom{0}78.10 / 250.60 & 76.474 & 78.544 \\
 & TWEO & 81.234 & \phantom{0}25.17 / \phantom{0}18.56 & 73.860 & 76.414 \\
 & CD & \textbf{81.628} & \phantom{0}45.58 / 162.44 & \textbf{77.784} & \textbf{78.576} \\
\midrule
\smash{\raisebox{-1.0\normalbaselineskip}{ViT-S}} & Baseline & 79.880 & \phantom{0}51.00 / 155.50 & 69.268 & 73.998 \\
 & TWEO & 80.176 & \phantom{0}29.10 / \phantom{0}59.21 & 68.552 & 74.472 \\
 & CD & \textbf{80.218} & \phantom{0}30.89 / 100.09 & \textbf{70.326} & \textbf{74.774} \\
\midrule
\smash{\raisebox{-0.5\normalbaselineskip}{ViT-T}} & Baseline & 72.720 & \phantom{0}22.29 / \phantom{0}32.71 & 53.910 & 62.606 \\
 & CD & \textbf{73.052} & \phantom{0}17.89 / \phantom{0}26.58 & \textbf{55.740} & \textbf{63.066} \\
\midrule
\smash{\raisebox{-1.0\normalbaselineskip}{Swin-T}} & Baseline & 81.092 & 277.95 / 498.34 & 71.286 & 79.430 \\
 & TWEO & 81.152 & \phantom{0}43.68 / \phantom{0}47.52 & 76.258 & 79.066 \\
 & CD & \textbf{81.252} & 122.08 / 139.93 & \textbf{76.484} & \textbf{79.682} \\
\bottomrule
\end{tabular}
\end{table}

Unless otherwise noted, all models are trained with AMP~\citep{micikevicius2017amp}, use the AdamW~\citep{loshchilov2017adamw} optimizer, and all quantization use 4-bit weights and 4-bit activations(W4A4). When CD is enabled, we reduce the original weight-decay coefficient so that $\lambda_{\mathrm{wd}} + \lambda_{\mathrm{cd}}$ matches the baseline regularization budget. Unless stated otherwise, we set $\lambda_{\mathrm{cd}}$ to one tenth of the original weight-decay coefficient. For APHQ~\citep{wu2025aphq}, we disable MLP reconstruction and keep block reconstruction, in order to reduce quantization time.

Besides task accuracy, we also report maximum activation. This quantity is measured on the validation set, aggregated either over the internal outputs of each modules and over the Transformer block outputs, or over the backbone and detector head outputs in detection. We report it as a diagnostic quantity rather than as a standalone target: lower values often help quantization, but aggressively suppressing them may also hurt FP or even quantization accuracy.

\subsection{ImageNet-1K Pre-training}

We follow the official DeiT~\citep{touvron2021deit} and Swin~\citep{liu2021swin} training recipes for ViT and Swin backbones. Table~\ref{tab:exp_pretrain_main} compares the baseline, CD, and TWEO on standard ImageNet-1K pre-training. For TWEO~\citep{liang2025tweo}, we reproduce the method in the same training framework using the method-specific hyperparameters provided by the authors, while keeping the base training hyperparameters aligned with the baseline and CD runs. A brief summary of TWEO is given in Appendix~\ref{app:tweo}.

CD consistently improves quantized accuracy over the matched baseline, while preserving or slightly improving FP accuracy. At the same time, it reduces maximum activation by a clear margin on the matched runs. The TWEO references further support the central motivation for CD: aggressively driving activations to very small values is not necessary for good deployment-time behavior, and in our cases it leads to a weaker FP/quantized trade-off than CD.

We also test whether CD is complementary to architectural outlier mitigation by applying it to three attention-sink variants~\citep{kaul2024softmax1,sun2024massive,agarwal2025attnbias} of ViT-B; the detailed setup and full table are deferred to Appendix~\ref{app:attn-sink}. These variants change the attention computation, whereas CD leaves the architecture unchanged and regularizes matrix-pair alignment. CD improves RepQ-ViT accuracy in all three cases. Combined with \texttt{softmax1}, it reaches the strongest RepQ-ViT result in ViT-B series, at 78.550.

\subsection{Efficiency}

One practical advantage of CD is that it is implemented as a decoupled weight update rather than as an extra loss term. Figure~\ref{fig:exp_efficiency} quantifies this difference on ImageNet-1K pre-training. We fix the global batch size to 1024, run five epochs, and average over three runs on Ampere GPUs.

The left panel shows that CD adds only a small overhead in both time and memory, and its extra memory is nearly constant. By contrast, TWEO~\citep{liang2025tweo} incurs a much larger time and memory overhead that grows linearly with batch size. The right panel shows the same pattern across hidden sizes at matched batch size: CD remains efficient, whereas TWEO becomes more expensive on the wider backbone. This matches the implementation difference between a direct decoupled matrix update and an activation-regularization loss that must flow through the training graph.

\subsection{COCO Detection}

We next study joint detector training in MMDetection~\citep{chen2019mmdetection} with Mask R-CNN~\citep{he2017maskrcnn} and Swin~\citep{liu2021swin} backbones. We initialize the backbone from official Swin checkpoints and train the backbone and detector jointly. CD, TWEO~\citep{liang2025tweo}, and the Noise~\citep{baskin2021noise} (uniform-noise injection in weights) baseline are applied only to the backbone. For TWEO, the original auxiliary loss is too unstable in this setting, so we rescale it to match the task-loss magnitude and tune $\lambda_{\mathrm{TWEO}}$ into the $10^{-4}$ range. This produces a strong task-specific baseline.

\begin{table}[t]
\caption{Detection results on COCO. We report $\mathrm{mAP}^{\mathrm{box}} / \mathrm{mAP}^{\mathrm{mask}}$. All quantization results use W4A4 settings.}
\label{tab:exp_det_compare}
\label{tab:exp_det_configs}
\centering
{\footnotesize
\textbf{(a) Method comparison on Swin-T 1x.}\\
\vspace{0.25em}
\setlength{\tabcolsep}{5pt}
\begin{tabular}{lccc}
\toprule
Method & Max. Act. (backbone/head) & FP & RepQ-ViT \\
\midrule
Baseline & 191.93 / \phantom{0}97.52 & 0.426 / 0.393 & 0.311 / 0.302 \\
Noise & 224.66 / \phantom{0}96.60 & 0.425 / 0.392 & 0.328 / 0.318 \\
TWEO & \phantom{0}15.27 / \phantom{0}99.19 & 0.410 / 0.379 & 0.341 / 0.326 \\
CD & 149.17 / \phantom{0}92.93 & \textbf{0.429 / 0.396} & \textbf{0.344 / 0.330} \\
\bottomrule
\end{tabular}
}

\vspace{0.70em}

{\footnotesize
\textbf{(b) Results across schedules, backbones, and quantization algorithms.}
\vspace{0.25em}
\setlength{\tabcolsep}{3pt}
\begin{tabular}{lcccccc}
\toprule
 & \multicolumn{2}{c}{Swin-T 1x} & \multicolumn{2}{c}{Swin-T 3x} & \multicolumn{2}{c}{Swin-S 3x} \\
\cmidrule(lr){2-3}\cmidrule(lr){4-5}\cmidrule(lr){6-7}
Metric & Base & CD & Base & CD & Base & CD \\
\midrule
FP mAP & 0.426 / 0.393 & \textbf{0.429 / 0.396} & 0.459 / 0.415 & \textbf{0.460 / 0.416} & 0.483 / 0.432 & \textbf{0.484 / 0.432} \\
Percentile & 0.259 / 0.256 & \textbf{0.304 / 0.296} & 0.326 / 0.319 & \textbf{0.336 / 0.324} & 0.397 / 0.376 & \textbf{0.407 / 0.381} \\
RepQ-ViT & 0.311 / 0.302 & \textbf{0.344 / 0.330} & 0.366 / 0.356 & \textbf{0.375 / 0.361} & 0.414 / 0.389 & \textbf{0.424 / 0.396} \\
APHQ & 0.379 / 0.359 & \textbf{0.381 / 0.362} & 0.390 / 0.380 & \textbf{0.402 / 0.385} & 0.406 / 0.392 & \textbf{0.425 / 0.404} \\
GPLQ & 0.395 / 0.368 & \textbf{0.399 / 0.373} & 0.417 / 0.390 & \textbf{0.421 / 0.391} & 0.445 / 0.407 & \textbf{0.449 / 0.409} \\
\bottomrule
\end{tabular}
}
\end{table}

CD improves the quantized detector substantially, while preserving and slightly improving the FP detector. The uniform-noise and TWEO baselines also help quantization, but both incur noticeable FP accuracy. This is the same qualitative pattern as in classification pre-training: suppressing activation extremes is useful, but the best deployment trade-off does not come from minimizing them as aggressively as possible.

Panel (b) of Table~\ref{tab:exp_det_configs} extends this comparison across training schedules, model sizes, and quantization algorithms. The gains remain visible for the simple percentile~\citep{banner2019percentile} PTQ baseline, for the reparameterization-based RepQ-ViT\citep{li2023repq}, for the reconstruction-based APHQ~\citep{wu2025aphq}, and for the lightweight QAT baseline GPLQ~\citep{liang2025gplq}. Additional notes comparing our reproduced checkpoints with official results are given in Appendix~\ref{app:exp-details}.

\subsection{Downstream Classification Fine-tuning}

We fine-tune the official \texttt{vit\_base.orig\_in21k} checkpoint from timm on nine datasets: five fine-grained benchmarks~\citep{maji2013fgvc}, SUN397~\citep{xiao2010sun397}, EuroSAT~\citep{helber2019eurosat} with the spatial split, SmallNORB~\citep{lecun2004smallnorb}, and ImageNet-1K~\citep{deng2009imagenet}. All methods share one common fine-tuning recipe across datasets, except that ImageNet-1K uses a larger batch size and more update steps. Across methods, we also keep the same base hyperparameters, aside from the additional method-specific coefficients required by CD and TWEO~\citep{liang2025tweo}. Because the fine-tuning schedule is shorter and the learning rate is smaller than in pre-training, we set $\lambda_{\mathrm{cd}}=0.01$ for all target matrices, except for $W_{F2}$ where we use $\lambda_{\mathrm{cd}}=0.02$, while keeping the same regularization-budget heuristic. TWEO again uses the stabilized loss rescaling described above. Additional fine-tuning details are provided in Appendix~\ref{app:additional-exp-details}.

\begin{table}[t]
\caption{Downstream fine-tuning. Each entry reports FP acc / RepQ-ViT acc / acc-drop. The average row is the simple average over all nine datasets. The Maximum Activation row is measured only on the ImageNet-1K fine-tuning run. Max. Act. reports module / block-output activation maxima.}
\label{tab:exp_finetune}
\centering
\footnotesize
\setlength{\tabcolsep}{4pt}
\begin{tabular}{lccc}
\toprule
Dataset & Baseline & CD & TWEO \\
\midrule
Flowers102 & 98.73 / 97.84 / \phantom{0}-0.88 & \textbf{98.82} / \textbf{98.43} / \textbf{\phantom{0}-0.39} & 98.72 / 98.33 / \textbf{\phantom{0}-0.39} \\
Pets & \textbf{95.52} / \textbf{94.16} / \phantom{0}-1.36 & 95.11 / 94.02 / \textbf{\phantom{0}-1.09} & 95.11 / 93.89 / \phantom{0}-1.22 \\
Aircraft & 72.73 / 68.35 / \phantom{0}-4.38 & \textbf{73.87} / \textbf{70.54} / \phantom{0}-3.33 & 72.58 / 69.73 / \textbf{\phantom{0}-2.85} \\
Cars & 90.89 / 88.20 / \phantom{0}-2.69 & \textbf{91.07} / \textbf{88.81} / \phantom{0}-2.26 & 89.30 / 88.56 / \textbf{\phantom{0}-0.73} \\
Food101 & \textbf{86.69} / 83.76 / \phantom{0}-2.93 & 86.35 / \textbf{84.83} / \textbf{\phantom{0}-1.52} & 85.91 / 83.83 / \phantom{0}-2.08 \\
SUN397 & 68.65 / 63.57 / \phantom{0}-5.08 & \textbf{68.67} / \textbf{66.41} / \textbf{\phantom{0}-2.26} & 67.09 / 64.08 / \phantom{0}-3.01 \\
EuroSAT & \textbf{97.56} / 96.92 / \phantom{0}-0.63 & \textbf{97.56} / \textbf{97.17} / \phantom{0}-0.39 & 97.30 / 96.93 / \textbf{\phantom{0}-0.37} \\
SmallNORB & 68.01 / 55.84 / -12.18 & 69.33 / 57.93 / \textbf{-11.40} & \textbf{70.53} / \textbf{58.43} / -12.10 \\
ImageNet-1K & \textbf{82.90} / 78.54 / \phantom{0}-4.36 & 82.70 / \textbf{79.24} / \textbf{\phantom{0}-3.46} & 82.57 / 78.91 / \phantom{0}-3.65 \\
\midrule
Average & 84.63 / 80.80 / \phantom{0}-3.83 & \textbf{84.83} / \textbf{81.93} / \textbf{\phantom{0}-2.90} & 84.35 / 81.41 / \phantom{0}-2.93 \\
Max. Act. & 824.12 / 1680.77 & 145.75 / 551.55 & 56.92 / 71.36 \\
\bottomrule
\end{tabular}
\end{table}

Across these nine fine-tuning tasks, CD gives the best average FP accuracy, the best average quantized accuracy, and the smallest average drop. Other methods may remain stronger on individual datasets, especially on some FP metrics, while CD provides the best overall trade-off across the full benchmark suite. This again supports the view that controlling activations to a reasonable range is more effective than pushing them as low as possible.

\subsection{Ablations}

We finally report three ablations on ViT-B pre-training. Table~\ref{tab:exp_ablation_cd}(a) varies the CD strength $\lambda_{\mathrm{CD}}$. A non-zero CD strength reduces activation outliers and improves post-training quantization over the baseline, with $\lambda_{\mathrm{CD}}=0.005$ giving the best RepQ-ViT accuracy. Table~\ref{tab:exp_ablation_cd}(b) compares the decoupled CD update with a loss-term variant. A direct loss-term implementation is unstable under matched strength. We therefore report stabilized CD-loss variants that first align the CD term to the scale of the main loss and then multiply it by a smaller coefficient. Even under this favorable setup, the decoupled update remains better in both FP and quantized accuracy. 

Table~\ref{tab:exp_ablation_pairs} studies which matrices receive CD. Applying CD only to the second matrix in composable pairs or only to the second matrix in functional pairs improves both FP and quantized accuracy. Combining these two target sets gives the best quantized result, whereas applying CD to both matrices in each functional pair degrades the quantized accuracy.

\begin{table}[t]
\caption{Ablations on CD strength and implementation form. Max. Act. reports module-level / block-output activation maxima.}
\label{tab:exp_ablation_cd}
\centering
{
\footnotesize
\begin{minipage}[t]{0.485\linewidth}
\centering
\textbf{(a) CD strength $\lambda_{\mathrm{CD}}$.}
\vspace{0.25em}
\setlength{\tabcolsep}{3pt}
\begin{tabular}{@{}l@{\hspace{10pt}}ccc@{}}
\toprule
\multicolumn{1}{c}{$\lambda_{\mathrm{CD}}$} & Max. Act. & FP & RepQ-ViT \\
\midrule
$0$ & 78.10 / 250.60 & 81.444 & 76.474 \\
$0.0025$ & 59.90 / 203.81 & \textbf{81.744} & 76.928 \\
$0.005$ & \textbf{45.58 / 162.44} & 81.628 & \textbf{77.784} \\
$0.01$ & 51.81 / 171.89 & 81.554 & 77.646 \\
\bottomrule
\end{tabular}
\end{minipage}
\hspace{0.006\linewidth}
\begin{minipage}[t]{0.485\linewidth}
\centering
\textbf{(b) Update form.}
\vspace{0.25em}
\setlength{\tabcolsep}{3pt}
\begin{tabular}{@{}l@{\hspace{10pt}}ccc@{}}
\toprule
Method & Max. Act. & FP & RepQ-ViT \\
\midrule
Baseline & 78.10 / 250.60 & 81.444 & 76.474 \\
Decay, 0.005 & \textbf{45.58 / 162.44} & \textbf{81.628} & \textbf{77.784} \\
Loss, 0.01 & 65.13 / 196.94 & 81.356 & 43.418 \\
Loss, 0.005 & 86.75 / 218.63 & 81.398 & 75.348 \\
\bottomrule
\end{tabular}
\end{minipage}
}

\end{table}

\begin{table}[t]
\caption{Ablation on which matrices receive CD. A decays the second matrix in each composable pair; B decays the second matrix in each functional pair; C decays both matrices in each functional pair. Max. Act. reports module-level / block-output activation maxima.}
\label{tab:exp_ablation_pairs}
\centering
\footnotesize
\setlength{\tabcolsep}{4pt}
\begin{tabular}{lccccc}
\toprule
Metric & Baseline & A: comp. second & B: func. second & C: func. both & A$\,{+}\,$B \\
\midrule
Max. Act. & 78.10 / 250.60 & 89.08 / 236.28 & 47.80 / 167.07 & 64.98 / 191.97 & \textbf{45.58 / 162.44} \\
FP Top-1 & 81.444 & \textbf{81.892} & 81.460 & 81.428 & 81.628 \\
RepQ-ViT & 76.474 & 76.830 & 76.804 & 76.062 & \textbf{77.784} \\
\bottomrule
\end{tabular}
\end{table}

\section{Conclusion}

We studied quantization-friendly training for vision Transformers through the lens of cross-matrix colinearity, and introduced Colinearity-Decay as a decoupled regularizer for ordered matrix pairs. The main finding was that activation outliers should be controlled rather than simply minimized: CD reduced the pairwise alignments that contributed to large activations, while preserving full-precision accuracy under AMP training. Its decoupled form made the method non-invasive to the model architecture, stable to apply, inexpensive in training overhead and orthogonal to existing outlier-mitigation methods. Across ImageNet-1K pre-training, COCO joint detector training, and transfer fine-tuning, CD consistently improved quantized accuracy under diverse quantization pipelines, while maintaining or slightly improving full-precision accuracy. These results suggest that structural regularization of Transformer matrix pairs is a practical way to prepare vision models for low-bit deployment without changing the inference architecture.

\section{Limitations and Future Work}

\begin{itemize}
    \item \textbf{Integration with QAT.} This work mainly applies CD before downstream quantization. Many deployment recipes include quantization-aware training or lightweight adaptation~\citep{liang2025gplq}, and combining CD with these objectives remains an open direction.
    \item \textbf{Broader architectures and tasks.} CD is defined on ordered matrix pairs and is not specific to ViT or Swin. Appendix~\ref{app:gpt2} provides an initial GPT-2~\citep{radford2019gpt2} language-modeling study, but broader validation on larger and more modern LLMs~\citep{grattafiori2024llama3,yang2025qwen3,liu2024deepseekv3}, diffusion Transformers~\citep{peebles2023dit}, and multimodal models~\citep{bai2025qwen3vl} may require revisiting the specific design.
    \item \textbf{Mechanistic scope.} CD targets cross-matrix alignment as one structural source of activation outliers, but Transformer outliers can also arise from attention dynamics, residual pathways, normalization behavior, and data-dependent routing. Future work should separate these sources more explicitly and study when pairwise matrix regularization is sufficient, and when it should be combined with architecture-level or activation-level interventions.
    \item \textbf{Configuration and scaling.} We use a simple fixed pair set, normalization factor, and regularization-budget heuristic. Future work should study better tuning strategies, layer-wise or scheduled CD strengths, more bit-widths, and larger-scale training. Sensitivity-aware or layer-wise coefficients may reduce manual tuning.
\end{itemize}

\begin{ack}
This work was partly supported by the National Natural Science Foundation of China under Grant 62276123 and Fundamental and Interdisciplinary Disciplines Breakthrough Plan of the Ministry of Education of China (No. JYB2025XDXM118).

JW identified colinearity as a structural source of activation outliers. JT proposed the Colinearity-Decay formulation and developed it into a complete method under JW's guidance. JW and JT wrote the paper. GL provided detailed guidance on experiments and writing. PS contributed to part of the experiments and writing.
\end{ack}

\appendix
\section{Appendix}

\subsection{FFN Outliers and FC1-FC2 Colinearity}
\label{app:ffn-outlier-colinearity}

\begin{figure}[h]
\centering
\includegraphics[width=\linewidth]{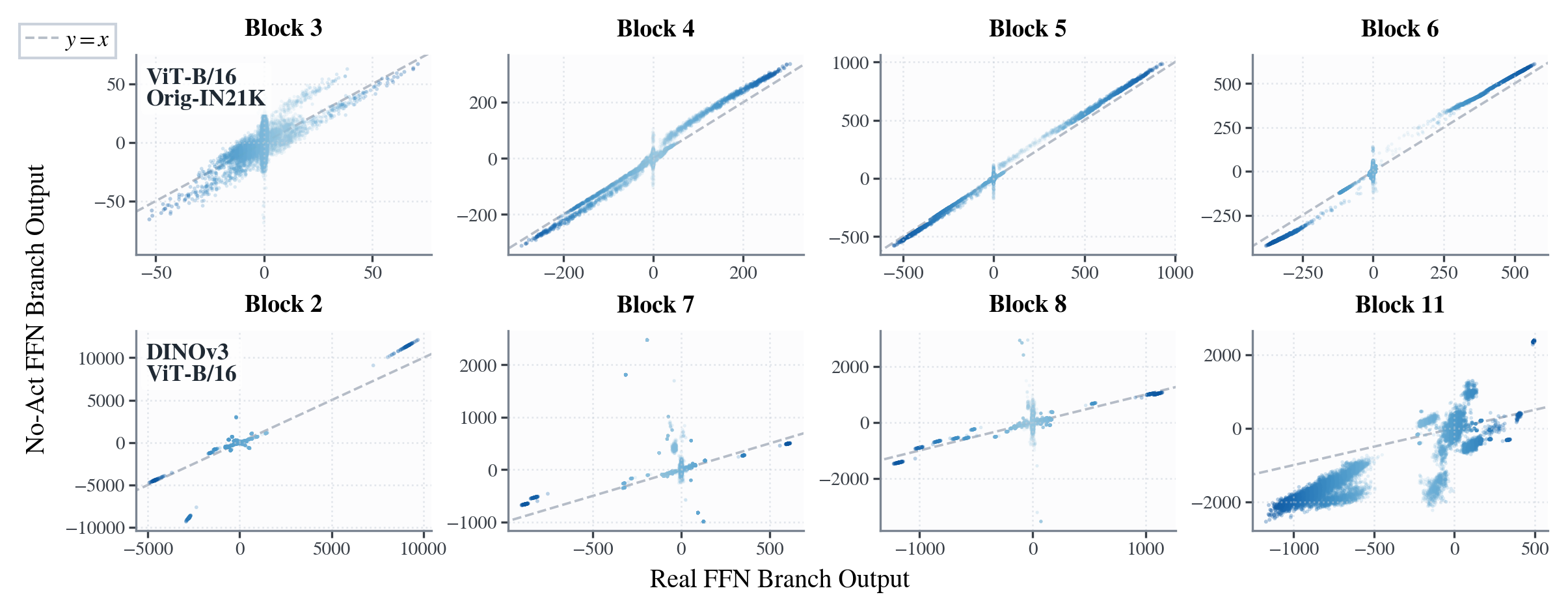}
\caption{Real FFN branch output versus the no-activation surrogate $W_{F2}W_{F1}x$ across selected blocks of ViT-B/16 and DINOv3 ViT-B/16. The dashed line denotes $y=x$.}
\label{fig:app_ffn_surrogate}
\end{figure}

We first examine whether the linear surrogate in Eq.~\eqref{eq:method_ffn_surrogate} captures the large coordinates produced by the actual FFN branch. Figure~\ref{fig:app_ffn_surrogate} compares the real FFN branch output with the no-activation approximation $W_{F2}W_{F1}x$ on selected blocks of ViT-B/16 and DINOv3~\citep{simeoni2025dinov3} ViT-B/16. Although the surrogate ignores the activation function, the large-magnitude outputs often follow the same directions as the real branch output. This agreement indicates that the matrix pair $(W_{F1}, W_{F2})$ already contains a structural source of FFN outliers.

Eq.~\eqref{eq:method_row_decomp} gives a more specific explanation of this effect. Let $W_{F1}=USV^\top$, and let $\hat{w}_{2,j}$ denote the $j$-th row of $W_{F2}$ after row normalization. We measure the alignment between the $i$-th left singular direction of $W_{F1}$ and the rows of $W_{F2}$ by
\begin{equation}
\alpha_{i,j} = |\hat{w}_{2,j}^{\top} u_i|.
\label{eq:app_ffn_alignment}
\end{equation}

\begin{figure}[t]
\centering
\includegraphics[width=\linewidth]{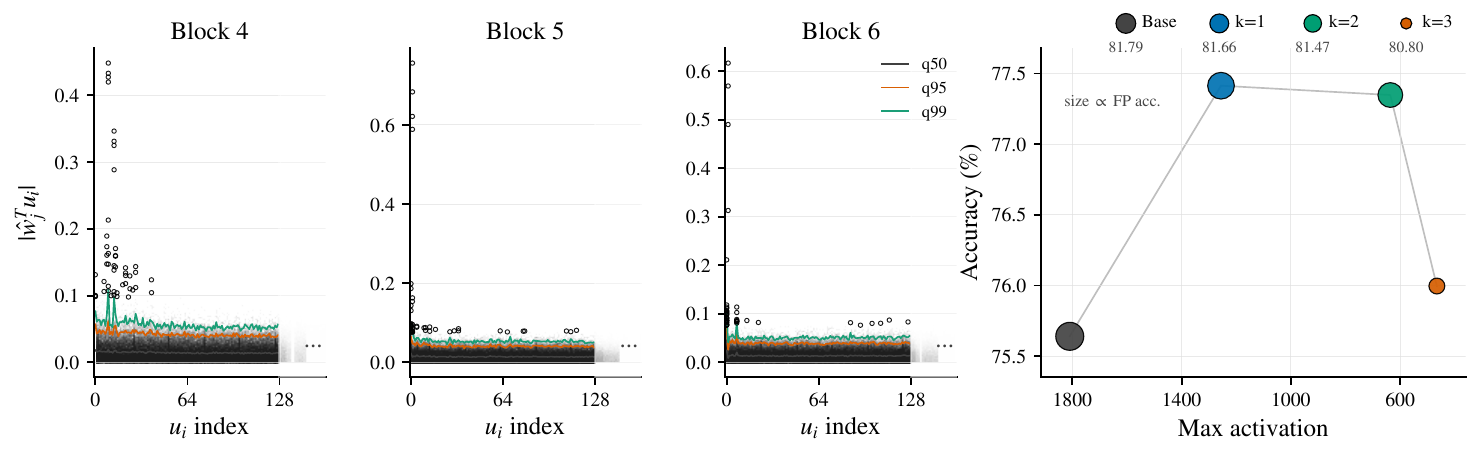}
\caption{FC1-FC2 alignment and its effect on quantization. Left: normalized alignment $|\hat{w}_{2,j}^{\top}u_i|$ between FC1 left singular vectors and FC2 rows in Blocks 4, 5, and 6. Right: effect of sequentially removing the most aligned singular directions on maximum activation and RepQ-ViT W4A4 accuracy. Marker area is proportional to FP accuracy. All conducted on \texttt{vit\_base.orig\_in21k\_ft\_in1k}}
\label{fig:app_ffn_alignment}
\end{figure}

Figure~\ref{fig:app_ffn_alignment} visualizes this alignment in outlier-heavy layers. Blocks 4, 5, and 6 contain a small number of singular-vector directions whose maximum row alignment is far above the bulk of the distribution. These peaks are not spread uniformly across all directions; they concentrate on a few FC1 left singular vectors. This pattern is consistent with the colinearity mechanism in Eq.~\eqref{eq:method_row_decomp}: the outlier is induced by a specific cross-matrix alignment rather than by a uniformly large FC2 row norm.

To test whether these directions are redundant, we aggregate the alignment scores over the selected layers, sort the strongest directions, and sequentially set the corresponding FC1 singular values to zero. The three removed directions are Block 5 index 1, Block 6 index 1, and Block 4 index 9. Removing the top one or two directions substantially reduces the maximum activation and improves RepQ-ViT~\citep{li2023repq} W4A4 accuracy while leaving FP accuracy close to the baseline.

This diagnostic intervention supports two conclusions. First, FFN outliers in ViT are tightly connected to the colinearity between dominant FC1 singular directions and selected FC2 rows. Second, part of this colinearity is redundant for the full-precision model. Controlling this cross-matrix alignment can therefore reduce activation outliers and improve the quantized model, which is the mechanism targeted by CD.

\subsection{Method Derivations and Details}
\label{app:cd-details}

We collect the derivations and implementation details omitted from the main text.

\textbf{CD gradient derivation.}
For the CD objective,
\begin{equation}
l_{\mathrm{cd}}(W_1, W_2)=\|W_2W_1\|_F^2
=\mathrm{tr}(W_1^\top W_2^\top W_2W_1),
\end{equation}
where we use $\|A\|_F^2=\mathrm{tr}(A^\top A)$ and the cyclic property of the trace. We compute gradients through differentials and use the convention
$\mathrm{d}l=\mathrm{tr}((\partial l/\partial W)^\top \mathrm{d}W)$.
First fix $W_2$ and set $A=W_2^\top W_2$. Since $A=A^\top$,
\begin{align}
\mathrm{d}_{W_1}l_{\mathrm{cd}}
&= \mathrm{tr}(\mathrm{d}W_1^\top A W_1)
 + \mathrm{tr}(W_1^\top A\,\mathrm{d}W_1) \nonumber \\
&= \mathrm{tr}((A W_1)^\top \mathrm{d}W_1)
 + \mathrm{tr}((A^\top W_1)^\top \mathrm{d}W_1) \nonumber \\
&= \mathrm{tr}((2A W_1)^\top \mathrm{d}W_1).
\end{align}
Thus $\partial l_{\mathrm{cd}}/\partial W_1=2W_2^\top W_2W_1$. Next fix $W_1$ and set $B=W_1W_1^\top$. By cyclicity,
$l_{\mathrm{cd}}=\mathrm{tr}(W_2 B W_2^\top)$, with $B=B^\top$. Therefore,
\begin{align}
\mathrm{d}_{W_2}l_{\mathrm{cd}}
&= \mathrm{tr}(\mathrm{d}W_2 B W_2^\top)
 + \mathrm{tr}(W_2 B\,\mathrm{d}W_2^\top) \nonumber \\
&= \mathrm{tr}((W_2 B^\top)^\top \mathrm{d}W_2)
 + \mathrm{tr}((W_2 B)^\top \mathrm{d}W_2) \nonumber \\
&= \mathrm{tr}((2W_2B)^\top \mathrm{d}W_2).
\end{align}
The gradients are therefore
\begin{equation}
\frac{\partial l_{\mathrm{cd}}}{\partial W_1}=2W_2^\top W_2 W_1,
\qquad
\frac{\partial l_{\mathrm{cd}}}{\partial W_2}=2W_2W_1W_1^\top,
\end{equation}
which matches Eq.~\eqref{eq:method_cd_grads}. Using $W_1=USV^\top$ gives $W_1W_1^\top=US^2U^\top$, so the $j$-th downstream row satisfies Eq.~\eqref{eq:method_cd_row_grad}. This shows that CD removes the component of $w_{2,j}$ in the left singular subspace of $W_1$, with larger shrinkage on directions with larger singular values.

\textbf{Directional compatibility with weight decay.}
We next show that, for a downstream row vector, the CD gradient and the weight-decay gradient form an angle no larger than $90^\circ$. Let
\begin{equation}
M = U S^2 U^\top \succeq 0,\qquad
g_{\mathrm{cd},j} = 2 w_{2,j}^\top M,\qquad
g_{\mathrm{wd},j} = w_{2,j}^\top,
\end{equation}
where $g_{\mathrm{cd},j}$ is the CD gradient in Eq.~\eqref{eq:method_cd_row_grad}, and $g_{\mathrm{wd},j}$ is the row-wise gradient of the weight-decay term $\frac{1}{2}\|W_2\|_F^2$. The matrix $M$ is positive semi-definite because, for any vector $z$,
\begin{equation}
z^\top M z = z^\top U S^2 U^\top z = \|S U^\top z\|_2^2 \geq 0.
\end{equation}
Using the Euclidean inner product between row gradients, we obtain
\begin{equation}
\langle g_{\mathrm{cd},j}, g_{\mathrm{wd},j}\rangle
= 2 w_{2,j}^\top M w_{2,j} \geq 0.
\end{equation}
When both gradients are nonzero, their cosine similarity is therefore nonnegative:
\begin{equation}
\cos\theta_j
=\frac{\langle g_{\mathrm{cd},j}, g_{\mathrm{wd},j}\rangle}
{\|g_{\mathrm{cd},j}\|_2\|g_{\mathrm{wd},j}\|_2}
\geq 0.
\end{equation}
Hence $\theta_j\in[0^\circ,90^\circ]$. If either gradient is zero, the angle is not defined, but CD still has no component opposing the weight-decay gradient.

\textbf{Normalized CD and regularization budget.}
The implementation uses two scale-control details. First, when CD is enabled, we reduce the original weight-decay coefficient by the CD coefficient so that the decay budget is approximately fixed. For example, if the baseline uses $\lambda_{\mathrm{wd}}=0.05$, we usually set $\lambda_{\mathrm{cd}}$ to one tenth of it, namely $\lambda_{\mathrm{cd}}=0.005$, and use $\lambda_{\mathrm{wd}}=0.045$ for the CD run.

Second, we normalize the upstream matrix in the CD objective. Let
\begin{equation}
\bar{W}_1
= \frac{W_1}{\|W_1\|_F / \sqrt{d_{\mathrm{out}}(W_1)}} .
\end{equation}
The implemented CD objective is
\begin{equation}
l_{\mathrm{cd}}^{\mathrm{norm}}(W_1,W_2)
= \|W_2\bar{W}_1\|_F^2
= \left\|W_2
\frac{W_1}{\|W_1\|_F / \sqrt{d_{\mathrm{out}}(W_1)}}
\right\|_F^2 .
\label{eq:app_cd_norm_objective}
\end{equation}
Because CD is applied only to the downstream matrix, the required derivative is the derivative with respect to $W_2$ with $W_1$ fixed:
\begin{equation}
\frac{\partial l_{\mathrm{cd}}^{\mathrm{norm}}}{\partial W_2}
= 2W_2\bar{W}_1\bar{W}_1^\top
= \frac{2d_{\mathrm{out}}(W_1)}{\|W_1\|_F^2}
W_2W_1W_1^\top .
\label{eq:app_cd_norm_w2_grad}
\end{equation}
As in the main text, the factor $2$ is absorbed into $\lambda_{\mathrm{cd}}$ in the decoupled update. The normalization removes most of the dependence of the CD step on the absolute Frobenius scale of $W_1$, while the budget adjustment avoids simply adding an extra decay source on top of the baseline. Together, these choices keep the baseline and CD models at roughly similar per-matrix Frobenius norms and control the comparison variable more cleanly.

\subsection{TWEO Background}
\label{app:tweo}

TWEO~\cite{liang2025tweo} is an activation-regularization method designed to suppress extreme activation outliers during training. Unlike CD, which is implemented as a decoupled update on matrix pairs, TWEO regularizes block-output activations directly through the training graph.

\begin{equation}
L_{\mathrm{total}} = L_{\mathrm{task}} + \lambda(t) L_{\mathrm{TWEO}},
\end{equation}
\begin{equation}
L_{\mathrm{TWEO}} = \frac{1}{L} \sum_{l=1}^{L} \mathbb{E} \left[ \left( \frac{|A^{(l)}|}{\tau + \epsilon} \right)^p \right],
\end{equation}
where $A^{(l)}$ denotes the final output activation of the $l$-th Transformer block, $\lambda(t)$ is the step-dependent loss weight, $\tau$ is a magnitude scale, and $p$ controls how strongly large activations are penalized. In the original paper, TWEO is applied to every block output, typically with $p=4$, $\tau=3$, and an optional cosine schedule on $\lambda(t)$.

When we applied the original TWEO formulation directly to vision classification fine-tuning and detector joint training, optimization often collapsed. For these settings, we therefore rescaled the auxiliary term to match the magnitude of the main task loss and tuned the TWEO hyperparameters for each task.

\subsection{Attention-Sink Variants}
\label{app:attn-sink}

We evaluate three attention-sink mitigation variants on ViT-B, all trained with the same ImageNet-1K recipe as the standard ViT-B runs in Table~\ref{tab:exp_pretrain_main}. The \texttt{kv-bias}~\citep{sun2024massive} variant appends one learnable vector after each of $K$ and $V$. The \texttt{attn-bias}~\citep{agarwal2025attnbias} variant adds one learnable bias term to the softmax denominator of each head. The \texttt{softmax1}~\citep{kaul2024softmax1} variant replaces the standard softmax denominator by a denominator with an additional constant $1$.

Table~\ref{tab:exp_sink_variants} shows that CD remains effective on all three variants. The simple percentile PTQ~\citep{banner2019percentile} baseline can become fragile on these architectures, but the RepQ-ViT results remain consistently improved. The strongest result in this group is obtained by combining CD with \texttt{softmax1}, which reaches 78.550 RepQ-ViT~\citep{li2023repq} top-1 on ViT-B.

\begin{table}[t]
\caption{Attention-sink mitigation variants on ViT-B. All models use the same ImageNet-1K training recipe as the standard ViT-B runs.}
\label{tab:exp_sink_variants}
\centering
\small
\setlength{\tabcolsep}{5pt}
\begin{tabular}{llcccc}
\toprule
Variant & Method & FP Top-1 & Max. Act. (mod./blk.) & Percentile & RepQ-ViT \\
\midrule
\smash{\raisebox{-0.5\normalbaselineskip}{ViT-B}} & Baseline & 81.444 & \phantom{0}78.10 / 250.60 & 59.862 & 76.474 \\
 & CD & \textbf{81.628} & \phantom{0}45.58 / 162.44 & \textbf{63.474} & \textbf{77.784} \\
\midrule
\smash{\raisebox{-0.5\normalbaselineskip}{\texttt{kv-bias}}} & Baseline & 81.710 & \phantom{0}88.32 / 139.85 & 26.886 & 75.132 \\
 & CD & \textbf{81.846} & \phantom{0}81.65 / 109.65 & \textbf{27.010} & \textbf{77.662} \\
\midrule
\smash{\raisebox{-0.5\normalbaselineskip}{\texttt{attn-bias}}} & Baseline & \textbf{81.776} & 138.50 / 207.71 & 60.916 & 76.034 \\
 & CD & 81.668 & \phantom{0}92.56 / 125.20 & \textbf{72.520} & \textbf{77.924} \\
\midrule
\smash{\raisebox{-0.5\normalbaselineskip}{\texttt{softmax1}}} & Baseline & \textbf{82.000} & 129.35 / 205.09 & 71.182 & 70.568 \\
 & CD & 81.740 & 110.11 / 151.21 & \textbf{74.618} & \textbf{78.550} \\
\bottomrule
\end{tabular}
\end{table}

\subsection{Official vs. Reproduced Detection Results}
\label{app:exp-details}

For detection, our reproduced checkpoints are trained and evaluated with MMDetection 3.3.0, PyTorch 2.2.1, and CUDA 12.1. We initialize the detector from the official Swin~\citep{liu2021swin} checkpoints and use the official MMDetection training configurations. Table~\ref{tab:app_det_reproduce} compares published official results with our reproduced baselines on the settings used in the main experiments. The official RepQ-ViT~\citep{li2023repq} and GPLQ~\citep{liang2025gplq} numbers are taken from GPLQ, while the official APHQ~\citep{wu2025aphq} numbers are taken from APHQ. APHQ does not report the Swin-T 1x setting, so that entry is marked with a dash.

\begin{table}[t]
\caption{Official versus reproduced quantized detection results. We report $\mathrm{mAP}^{\mathrm{box}} / \mathrm{mAP}^{\mathrm{mask}}$, and $\Delta$ is computed as reproduced minus official.}
\label{tab:app_det_reproduce}
\centering
\small
\setlength{\tabcolsep}{4pt}
\begin{tabular}{llccc}
\toprule
Model & Source & RepQ-ViT & APHQ & GPLQ \\
\midrule
\smash{\raisebox{-1.0\normalbaselineskip}{Swin-T 1x}} & Official & \phantom{+}0.135 / \phantom{+}0.137 & --- & \phantom{+}0.379 / \phantom{+}0.368 \\
 & Reproduced & \phantom{+}0.311 / \phantom{+}0.302 & \phantom{+}0.379 / \phantom{+}0.359 & \phantom{+}0.395 / \phantom{+}0.368 \\
 & $\Delta$ & \phantom{+}\llap{+}0.176 / \phantom{+}\llap{+}0.165 & --- & \phantom{+}\llap{+}0.016 / \phantom{+}\llap{+}0.000 \\
\midrule
\smash{\raisebox{-1.0\normalbaselineskip}{Swin-T 3x}} & Official & \phantom{+}0.361 / \phantom{+}0.360 & \phantom{+}0.389 / \phantom{+}0.381 & \phantom{+}0.401 / \phantom{+}0.389 \\
 & Reproduced & \phantom{+}0.366 / \phantom{+}0.356 & \phantom{+}0.390 / \phantom{+}0.380 & \phantom{+}0.417 / \phantom{+}0.390 \\
 & $\Delta$ & \phantom{+}\llap{+}0.005 / \phantom{+}\llap{-}0.004 & \phantom{+}\llap{+}0.001 / \phantom{+}\llap{-}0.001 & \phantom{+}\llap{+}0.016 / \phantom{+}\llap{+}0.001 \\
\midrule
\smash{\raisebox{-1.0\normalbaselineskip}{Swin-S 3x}} & Official & \phantom{+}0.426 / \phantom{+}0.400 & \phantom{+}0.441 / \phantom{+}0.410 & \phantom{+}0.434 / \phantom{+}0.413 \\
 & Reproduced & \phantom{+}0.414 / \phantom{+}0.389 & \phantom{+}0.406 / \phantom{+}0.392 & \phantom{+}0.445 / \phantom{+}0.407 \\
 & $\Delta$ & \phantom{+}\llap{-}0.012 / \phantom{+}\llap{-}0.011 & \phantom{+}\llap{-}0.035 / \phantom{+}\llap{-}0.018 & \phantom{+}\llap{+}0.011 / \phantom{+}\llap{-}0.006 \\
\bottomrule
\end{tabular}
\end{table}

The reproduced baselines do not exactly match the official numbers. Some settings are higher in our reproduction, especially Swin-T 1x under RepQ-ViT and the GPLQ results on Swin-T, while others are lower, most clearly Swin-S under RepQ-ViT and APHQ. We suspect that these discrepancies come mainly from the newer software stack and different hardware generation. They do not affect the main conclusion: in the controlled comparisons in Table~\ref{tab:exp_det_configs}, CD consistently improves over the reproduced baseline across the detection settings and quantization algorithms.

\subsection{GPT-2 Language Modeling}
\label{app:gpt2}

Although the main experiments focus on vision Transformers, CD is defined on ordered matrix pairs in Transformer blocks and does not depend on image-specific operations. We therefore include a small language-modeling study on GPT-2~\citep{radford2019gpt2}. The experiments use nanoGPT and train naive GPT-2 models on OpenWebText~\citep{gokaslan2019openwebtext}. We report validation perplexity, where lower is better.

We use two model sizes. GPT-2 Small has 124M parameters, 12 Transformer layers, 12 attention heads, and hidden dimension 768. GPT-2 Medium has 350M parameters, 24 Transformer layers, 16 attention heads, and hidden dimension 1024. Both models use context length 1024, AdamW with weight decay 0.1 in the baseline, a global batch size of approximately 0.5M tokens, 300k training steps, and 1000 warmup steps. For GPT-2 Small, we follow the nanoGPT GPT-2 recipe and use peak learning rate $6.0{\times}10^{-4}$. For GPT-2 Medium, we use the common scale-up practice of reducing the peak learning rate to $3.0{\times}10^{-4}$ rather than treating this value as an official GPT-2 recipe.

For CD, we follow the regularization-budget rule used in the main experiments. GPT-2 Small uses $\lambda_{\mathrm{cd}}=0.005$ and reduces weight decay to 0.095. GPT-2 Medium uses $\lambda_{\mathrm{cd}}=0.01$ and reduces weight decay to 0.09. We evaluate bf16 models and three post-training quantization settings: RTN-W4A16, GPTQ-W4A16~\citep{frantar2022gptq}, and SmoothQuant-W8A8~\citep{xiao2023smoothquant}. For SmoothQuant, the non-strict setting uses per-token activation quantization. The strict setting uses per-tensor activation quantization and quantizes the attention-related batch-matrix multiplications related to attention operations.

\begin{table}[t]
\caption{GPT-2 language modeling on OpenWebText. We report validation perplexity; lower is better.}
\label{tab:app_gpt2}
\centering
\small
\setlength{\tabcolsep}{5pt}
\begin{tabular}{lcccc}
\toprule
 & \multicolumn{2}{c}{GPT-2 Small} & \multicolumn{2}{c}{GPT-2 Medium} \\
\cmidrule(lr){2-3}\cmidrule(lr){4-5}
Setting & Baseline & CD & Baseline & CD \\
\midrule
bf16 & 17.52 & \textbf{17.51} & 13.72 & \textbf{13.70} \\
RTN-W4A16 & 45.46 & \textbf{41.90} & 30.76 & \textbf{25.16} \\
GPTQ-W4A16 & \textbf{25.00} & 26.99 & 16.86 & \textbf{15.93} \\
SmoothQuant-W8A8 & 17.78 & \textbf{17.74} & 13.76 & \textbf{13.70} \\
SmoothQuant-W8A8-strict & 30.05 & \textbf{23.70} & 22.27 & \textbf{21.10} \\
\bottomrule
\end{tabular}
\end{table}

Table~\ref{tab:app_gpt2} shows that CD preserves the bf16 validation perplexity on both model sizes. Except for GPTQ-W4A16 on GPT-2 Small, all quantized settings improve over the corresponding baseline. The gains are especially clear under the stricter SmoothQuant setting, where activation quantization is more demanding. The W4A16 results test weight-only quantization, while the W8A8 results quantize both weights and activations. This pattern indicates that CD makes GPT-2 models friendlier to both weight quantization and activation quantization. These results provide initial evidence beyond vision Transformers, and they motivate evaluating CD on larger scales and more modern LLM architectures.

\subsection{Additional Experimental Details}
\label{app:additional-exp-details}

\textbf{Compute resources.}
Our experiments were run on nodes with NVIDIA A100 GPUs, Intel Xeon Platinum 8362 CPUs, and roughly 460 GiB of host memory. ImageNet-1K pre-training and COCO detection used distributed data parallel training with up to eight GPUs. Downstream fine-tuning experiments were run on a single GPU.

\textbf{Downstream fine-tuning recipe.}
For the eight non-ImageNet downstream datasets in Table~\ref{tab:exp_finetune}, we fine-tune \texttt{vit\_base\_patch16\_224.orig\_in21k} from the timm checkpoint with batch size 64 and evaluation batch size 256. We use AdamW, AMP training, weight decay 0.05, peak learning rate $5{\times}10^{-4}$, layer-wise learning-rate decay 0.9, drop-path rate 0.1, label smoothing 0.1, and a cosine learning-rate schedule with 5 warmup epochs followed by 75 cosine epochs. The data augmentation follows the timm loader defaults used by our script: RandAugment \texttt{rand-m9-mstd0.5-inc1}, random erasing probability 0.25 with pixel mode, horizontal flip probability 0.5, bicubic interpolation, and the model-specific mean, standard deviation, crop percentage, and crop mode.

For the ImageNet-1K downstream fine-tuning row, we use the same model family and optimizer choices, but run a step-based recipe with batch size 128 per GPU(4 GPUs total), evaluation batch size 1024, 500 warmup steps, 20k total steps.

\bibliographystyle{unsrtnat}
\bibliography{refs}

\end{document}